\documentclass{article} 
\usepackage{times}
\usepackage{times}
\usepackage{epsfig}
\usepackage{graphicx}
\usepackage{amsmath}
\usepackage{amssymb}
\usepackage{algorithmic}
\usepackage{algorithm}
\usepackage{color}
\usepackage{url}
\usepackage{pict2e}
\usepackage{graphicx}
\usepackage{hyperref}
\usepackage{icml2014} 
\definecolor{mydarkblue}{rgb}{0,0.08,0.45}
\hypersetup{ %
pdfauthor={},
pdfsubject={CVPR},
pdfkeywords={},
pdfborder=0 0 0,
pdfpagemode=UseNone,
colorlinks=true,
linkcolor=mydarkblue,
citecolor=mydarkblue,
filecolor=mydarkblue,
urlcolor=mydarkblue,
pdfview=FitH}

\def\singleletter#1{\rotatebox[origin=B]{90}{#1}}
\makeatletter
\def\parseletters#1{\@tfor \@tempa := #1 \do {\kern2pt\singleletter{\@tempa}}}
\makeatother

\begin{document}

\title{Sequentially Generated Instance-Dependent Image Representations for Classification}
\author{
\and Gabriel Dulac-Arnold\\
\and Ludovic Denoyer\\
\and Nicolas Thome\\
\and Matthieu Cord\\
\and Patrick Gallinari\\
\\
LIP6, UPMC - Sorbonne University\\
Paris, France\\
{\tt\small \{firstname.lastname\}@lip6.fr}
}

\maketitle


\begin{abstract}
 
In this paper, we investigate a new framework for image classification that adaptively generates spatial representations. Our strategy is based on a sequential process that learns to explore the different regions of any image in order to infer its category. In particular, the choice of regions is specific to each image, directed by the actual content of previously selected regions.The capacity of the system to handle incomplete image information as well as its adaptive region selection allow the system to perform well in budgeted classification tasks by exploiting a dynamicly generated representation of each image.
We demonstrate the system's abilities in a series of image-based exploration and classification tasks that highlight its learned exploration and inference abilities.
\end{abstract}

\section{Introduction} \label{sec:introduction}

Many computer vision models are developped with a specific image classification task in mind, adapted to a particular representation such as the bag-of-words (BoW) model or low-level local features~\cite{sivic03, DBLP:journals/pami/GemertVSG10}. In these representations, all the image information is used to take the decision, even in the spatial BoW extension of Lazebnik et al.~\cite{lazebnik2006beyond}.

However, as pointed out in recent work, humans do not need to consider the entire image to be able to interpret it~\cite{DBLP:conf/cvpr/SharmaJS12}. 
On the contrary, humans are able to rapidly pick out the important regions of an image necessary to interpret it.
This fact suggests that concentrating on specific subset of image regions in an intelligent manner should be sufficient to properly classify an image.
In addition to simply selecting regions of an image, our system can actively decide to consider certain regions of an image in more detail by increasing the BoW resolution for
specific sub-regions of an image. This allows the system to adaptively use more or less resources when classifying images of varying complexity. Similar performance-oriented
goals have been recently put forward by Karayev et al.~\cite{Karayev:2012:Timely}.

\begin{figure}[t]
\centerline{\includegraphics[width=\linewidth]{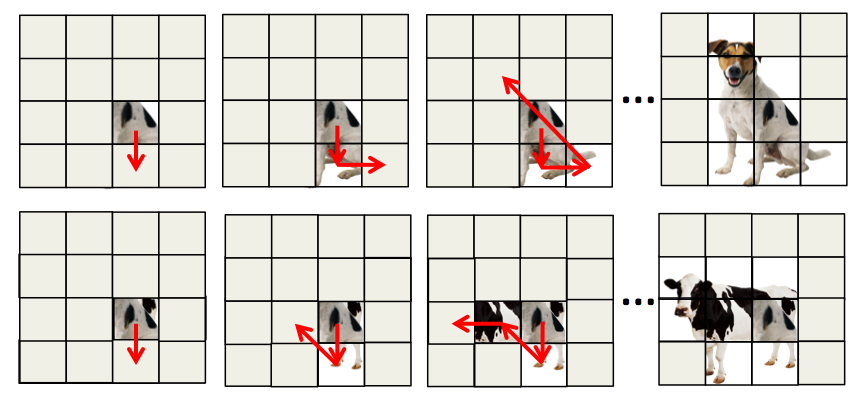}}
\caption{Illustration of our  classification framework for two test images (first and second line): 
According to the content of the center region, the next region to visit is selected (red arrow). Again, depending on the two first regions' contents, a third one is selected, and so on.  After $B$ iterations, the final classification is achieved.
As the first region is the same on both images, the second region explored is the same, but as these new regions' contents differs, the next regions considered by the algorithm are different.}
\vspace{-0.5cm}
\label{fig:toysmodel}
\end{figure} 

Importantly, this process is “instance-specific”, allowing the algorithm to adapt the choice of regions for each processed image. We are able to learn such a model by leveraging the datum-Wise classification framework~\cite{Dulac-Arnold2012b}, which is able to learn adaptive classification policies using reinforcement learning (RL).
We show that during inference, a significant speed-up is obtained by only computing the local features on the selected regions, while preserving acceptable inference accuracy w.r.t. full-information models.
The rest of the paper is organized as follows. Section 2 presents related works and highlights our contributions. Section 3 gives the theoretical background of our sequential model, and Section 4 details the training algorithm.
Finally, Section 5 reports classification scores on two challenging image datasets.

\section{Background} 
\label{sec:bg}
The standard image classification pipeline follows three steps~\cite{DBLP:conf/cvpr/BoureauBLP10} --- (i) low-level local descriptor extraction, (ii) coding, and (iii) pooling --- to  get feature vectors that are then used for classification.
This strategy was definitively popularized in computer vision with the bag-of-words formalism using SIFT local features~\cite{sivic03}.
Alternatives to the standard coding scheme have been  proposed, such as local soft coding~\cite{liu2011defense} or sparse coding~\cite{DBLP:conf/cvpr/BoureauBLP10}.
After the coding phase, most traditional BoW approaches use {\em sum pooling} or {\em max pooling}.  The spatial pyramid matching (SPM) strategy~\cite{lazebnik2006beyond} extends the pooling by considering a fixed predetermined spatial image pyramid. 

Many feature detectors have been proposed to get salient areas, affine regions, and points of interest~\cite{MikolajczykS05} on images. 
However, in contrast to the task of matching a specific target image or object, methods for category classification show better performance when using a uniform feature sampling over a dense grid on the image~\cite{chatfield11devil}.

Other approaches studying are motivated by human eye fixation or salient object detection \cite{IttiECCV2012,changICCV2011}.
Recently, several approaches combine dense sampling and saliency map or spatial weighting to obtain powerful image representations \cite{DBLP:journals/ijcv/SuJ12, DBLP:conf/cvpr/FengNTY11}. 
Sharma et al. \cite{DBLP:conf/cvpr/SharmaJS12} proposes a scheme to learn discriminative saliency maps at an image region level. 
They use the SPM scheme and apply weights to each block of the pyramid to get a global saliency map. 
In the case of multiclass classification, the weights are learned for each class in a discriminative way using one-against-all binary classification, and the final decision depends on the image content using a latent SVM representation.

Other methods based on latent SVM formulation also attempt to jointly encode spatial and content information in image classification. 
Parizi et al. ~\cite{DBLP:conf/cvpr/PariziOF12} introduce a reconfigurable model where each region is equiped with a latent variable representing a topic, such that only regions with similar topics are matched together in the final representation. 
This model provides a flexible framework, overcoming the shortcoming of the fixed spatial grid used in SPM. 

In all these approaches, the whole image has to be processed and all the information is used to classify, even if some regions of the image contain some misleading or irrelevant contents. 
We propose a strategy to overcome these limitations: we avoid processing the whole image by focusing only on the most pertinent regions relative to the image classification task.
In a way more drastic than Sharma's approach \cite{DBLP:conf/cvpr/SharmaJS12}, we constrain our system to take a decision by considering only a fixed number of regions of the test image, thus allowing the computation of the local features to be significantly reduced.

The most important aspect of our method is the region selection model.  In short, our model is effectively a learned sequential decision policy that sequentially chooses the best region to visit given a set of previously visited regions.  Both the locality and actual contents of a region are used in a joint manner to represent the set of visited regions.
Inspired by reinforcement learning algorithms, we propose a dedicated algorithm to learn the region selection policy used in our image classification task. 

More recent work uses similar techniques to find optimal orders for anytime object detection tasks~\cite{Karayev:2012:Timely}.  Similar work has been presented that uses a foveal glimpse simulation, but the learning approach is quite different~\cite{Larochelle2010}.

Sequential learning techniques have been recently applied to different standard classification tasks. 
In \cite{mlj}, the authors propose to use leinforcement learning models for learning sparse classifiers on vectors, \cite{kegl} use sequential techniques for learning a cascade of classifiers depending on the content of the inputs, 
while \cite{ecir} is an application of sequential learning models to text classification. Finally, \cite{thomas} propose a generic model able to minimize the data consumption with sequential online feature selection.

If our approach shares some common ideas with these recent works, we propose an original method that has been developed to handle the specific problem of 
classifying images using a small set of regions and a new learning algorithm which is efficient both in term of speed and performance. 

To summarize, the contributions presented in this paper are as follows:
\begin{itemize}
\item We propose a sequential model that, given an image, first selects a subset of relevant regions in this image, and then classifies it. The advantages of such a method are: 
(i) The classification decision is based only on the features of the acquired regions, resulting in a speed-up of the classification algorithm during inference. 
(ii) The algorithm is able to ignore misleading or irrelevant regions.
(iii) The way regions are selected depends both on the position but also on the content of the regions, resulting in a model able to adapt its behavior to the content of each image being classified. 
(iv) At last, the model is a multiclass model and the regions selection policy is learned globally for all the classes while other existing methods usually apply a category-specific region selection scheme.
\item We propose a new learning algorithm inspired from reinforcement learning techniques adapted to the particular problem faced here.
\item We present an experimental evaluation of this method on three different classical datasets and propose a qualitative study explaining the behaviour of this model.
\end{itemize}

\section{Classification model}
\label{metho}

\subsection{Notations}
Let us denote $\mathcal{X}$ the set of possible images and $\mathcal{Y}$ the discrete set of $C$ categories. A classifier is a parametrized function $f_\theta$ such that $f_\theta: \mathcal{X} \rightarrow \mathcal{Y}$ where $f_\theta(x)=y$ means that category\footnote{We consider in this paper the case of \textit{monolabel classification} where one input is associated to exactly one possible category.} $y$ has been predicted for image $x$. To learn $f_\theta$, a set of $\ell$ labeled training images $S_\text{train} = \{(x_1,y_1),...,(x_\ell,y_\ell)\}$ is provided to the system.

\begin{figure}
 \begin{center}
 \begin{tabular}{p{0.5\linewidth}p{0.5\linewidth}}
 \includegraphics[width=0.8\linewidth]{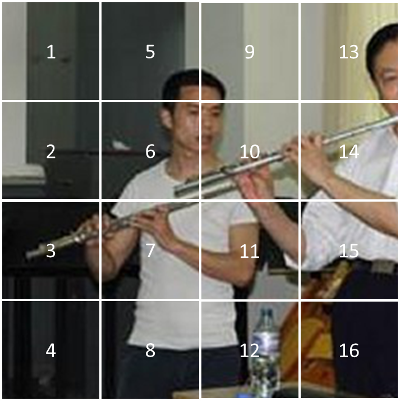} & \includegraphics[width=0.8\linewidth]{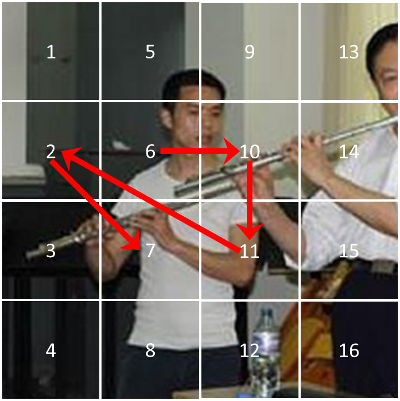}  \\
 \end{tabular}
 \end{center}
 \caption{(left) Index of the regions of an image decomposed of $4 \times 4$ regions. (right) Example of possible trajectory $(6,10,11,2,7)$.}
 \label{fig:i1}
 \end{figure}

We also consider for $x$ a fixed grid $N\times M$ of regions $\{r^x_i\}_{i\ \leq N \times M}$ where $r^x_i$ is the i-th region as illustrated in Fig.~\ref{fig:i1} (left). The set of all possible regions is denoted $\mathcal{R}$, and $\mathcal{R}(x)$  corresponds to the set of regions over image $x$.\footnote{Note that all the images have the same  $N\times M$ number of regions.}  $r^x_i$ is represented by a feature vector $\phi(r^x_i) $  of size $K$. We  use a SIFT bag-of-words representation in our experiments.

\subsection{Model formalization}
The classifier is modeled as a sequential decision process that, given an image,  first  sequentially selects regions, and then classifies the image using the information available in the visited regions. At each step, the classifier has already selected a sequence of regions denoted $(s^x_{1},..,s^x_{t})$ where $s^x_t$ is the index of the region of $x$ selected at step $t$. 
 The sequence $(s^x_{1},..,s^x_{t})$ is thus a representation tailored to the specific image and the current classification task.  $\mathcal{S}(x)$ denotes the set of all possible trajectories over image $x$ and $\mathcal{S}^t(x)$ the trajectories composed of $t$ selected regions.

Given a fixed budget $B$,  new regions are acquired resulting in a trajectory of size $B$. Given this trajectory, the classifier then decides which category to assign to the image. 
There are two important aspects of our approach: First, the way these regions are acquired depends on the content of the previously acquired regions --- c.f. Section \ref{sec:explo} --- resulting in a classifier that is able to adapt its representation to each image being classified, thus selecting the \textit{best regions} for each image. Second, the final decision is made given the features of the acquired regions only, without needing the computation of the features for the other regions, thus resulting in both a speed-up of the classification process --- not all features have to be computed --- but also, for some cases as described in Section \ref{sec:experiments}, in an improvement of the classification rate due to the exclusion of noisy regions. 

 We now give details concerning the \textbf{features}, the \textbf{classification phase} --- which classifies the image given the $B$ previously selected regions --- and the \textbf{exploration phase} --- which  selects $B-1$ additional regions\footnote{Note that we consider that the first region acquired by the classifier is a predefined central region of the image.} over an image to classify.

\subsubsection{Feature Function}
\label{sec:feat}
As previously explained, the $\Phi$ function  aims at aggregating the content of already visited regions. 
Based on the $K$ length vector $\phi(r^x_i)$, we consider the following $\Gamma$ mapping into a larger space of size $K \times (N \times M)$ as in \cite{DBLP:conf/cvpr/PariziOF12}: $\Gamma(\phi(r^x_i),K) = ( 0 \;  \dots   \;0  \; \phi(r^x_i) \; 0 \; \dots \; 0 )^T$
where $\phi(r^x_i)$ is positioned at index $i \times K$.  
The global feature function $\Phi$ is therefore defined as:
\begin{equation}
\Phi(s^x_{1},...s^x_{t}) = \sum\limits_{i=1}^t \Gamma(\phi(r^x_{s^x_{i}}),K).
\end{equation}
The goal of such a transformation is to conserve the information concerning the index of the region, which corresponds to the actual position of the region within the source image.

\subsubsection{Classification phase}

The classification phase consists in classifying an image given $B$ acquired regions denoted $(s^x_1,...s^x_{B})$. First, this set is transformed to a global feature vector that aggregates the individual features of each of its regions using $\Phi$. The classification is performed by using a classification function denoted $f_\theta$ defined as:
\begin{equation}
f_\theta : \begin{cases}
	\mathbb{R}^{K\times N \times M} \rightarrow \mathcal{Y}\\
	f_\theta(\Phi(s^x_{1},...s^x_{B})) = y
\end{cases},
\end{equation}
where $y$ is the predicted category. $\theta$ is the set of parameters that is learned using the training set as described in Section \ref{sec:learn}. Note that this function is computed by using as an input the vectorial representation of the sequence of regions $\Phi(s^x_{1},...s^x_{B})$. 

\subsubsection{Exploration phase}
\label{sec:explo}
In order to sequentially acquire the different regions of an image, the classification process follows an exploration policy denoted $\pi_\gamma$ where $\gamma$ is the set of parameters of this policy. A specific exploration policy is used at each timestep, and as such, $\pi$ can be decomposed into a sequence of sub-policies $\pi=(\pi^1,\pi^2, ...,\pi^{B-1})$ such that $\pi^t$ computes the region to acquire, at time $t$, given a sequence of $t$ previously acquired regions:
\begin{equation}
\forall t, \pi^t : \begin{cases} \mathbb{R}^{K\times N \times M} \rightarrow N \times M \\ \pi^t( \Phi(s^x_1,...,s^x_t) ) = s^x_{t+1}\end{cases},
\end{equation}
where $s^x_{t+1}$ is the index of the next region $r^x_{s^x_{t+1}}$ to acquire. $\pi^t$ can be viewed as a multiclass classifier which predicts the index of the next region to acquire, given the features of the previously acquired regions, and $\pi^t$ is restricted to predicting the index of a region that has not been previously acquired.  

In this paper, we  consider $t$ policies $\pi_{\gamma^t}$  parametrized by $\gamma^t$.  Similarly to the classification function $f_\theta$, $\pi_{\gamma^t}$ takes as an input the vectorial representation of the current sequence of acquired regions $\Phi(s^x_1,..,s^x_t)$ and outputs a region index to be considered.  We define this policy as a multiclass one-against-all hinge loss perceptron in this paper, but any multiclass classifier such as an SVM or a neural networks could be used as well.

\subsubsection{Final Inference Policy}

A complete classifier policy is defined by both an exploration policy $(\pi_1,...,\pi_{B-1})$ plus a classification policy $f_\theta$. The final inference process is described in Algorithm \ref{alg1} and consists in sequentially acquiring new regions (lines 1--4) and then computing the predicted category using the previously acquired regions (line 5).
 
\begin{algorithm}[t]
\begin{algorithmic}[1]
\REQUIRE $B$: budget
\REQUIRE $(\pi_1,...,\pi_{B-1})$: exploration policy
\REQUIRE $f_\theta$: classification policy
\REQUIRE $x$: input image
\STATE Acquire region $s^x_1$ i.e. the central region of the image
\FOR{$i=1$ \TO $B-1$}
\STATE Acquire region $s^x_{i+1}$ using $\pi_{\gamma^i}( \Phi(s^x_1,...,s^x_i))$
\ENDFOR
\STATE Compute category $y = f_\theta( \Phi(s^x_1,...,s^x_B))$
\RETURN $y$
\end{algorithmic}
\caption{Inference Algorithm: only $B$ regions are acquired for classification}
\label{alg1}
\end{algorithm}


\section{Learning Algorithm} 
\label{sec:learn}
The idea of the learning algorithm is the following: the classification policy is learned starting \textit{from the end}. We begin by first learning $f_\theta$ and then we sequentially learn $\pi_{\gamma^{B-1}}$, $\pi_{\gamma^{B-2}}$, up to $\pi_{\gamma^1}$. The underlying idea is to begin by learning a \textit{good} $f_\theta$ classification policy able to obtain good performance given any subset of $B$ regions. The learning of $\pi_{\gamma^{B-1}}$, $\pi_{\gamma^{B-2}}$, ... to $\pi_{\gamma^1}$  aims at acquiring relevant regions i.e. regions that will help $f_\theta$ to take the right decision.\linebreak

The complete learning algorithm is given in Algorithm \ref{alg2} and provides the general idea behind our method. At each iteration of the  algorithm, a set of learning states is sampled from the training images using a uniform random distribution. For each sampled state, the previously learned sub-policies are then used to simulate -- using Monte Carlo techniques -- the behavior of the algorithm and thus to provide supervision to the sub-policy we are learning. The detailed process is given in Sections \ref{sec:lc} and \ref{sec:ec}.  Note that this learning algorithm is original and derived from both the rollout classification policy iteration (RCPI) method and the Fitted-Q Learning model that are generic Reinforcement Learning models proposed in \cite{RCPI} and \cite{FQL}. Our method is an adaptation of these two classical algorithms in the particular case described here.

\begin{algorithm}[t]
\begin{algorithmic}[1]
\REQUIRE $(x_1,...,x_\ell)$: Training set of images
\REQUIRE $(y_1,...,y_\ell)$: Training labels
\STATE Learn $f_\theta$ using algorithm \ref{alg3}
\FOR{$k=B-1$ \TO $1$}
\STATE Use previoulsy learned sub-policies $\pi_\gamma^{k+1},.., \pi_\gamma^{B-1},f_\theta$ to learn $\pi_{\gamma^k}$ using algorithm \ref{alg4}
\ENDFOR
\RETURN Final policy: $(\pi_{\gamma^1},...\pi_{\gamma^{B-1}},f_\theta)$
\end{algorithmic}
\caption{Complete Learning algorithm}
\label{alg2}
\end{algorithm}

\subsection{Learning the Classification Policy}
\label{sec:lc}

Our approach for learning the classification policy is described in Algorithm \ref{alg3}.  
 The underlying idea is to automatically learn from the training set a classifier which is optimal for classifying \textit{any} subset of $B$ regions for \textit{any} input image. The classification policy $f_\theta$ is learned on a large training set of images which have had $B$ regions uniformly sampled --- lines 5 and 6 of Alg. \ref{alg3} --- over the training images. Each set of regions is transformed to a feature vector using the $\Phi$ function and provided to the learning algorithm using the label of the image as the supervision. This set of regions and labels is used to train $f_\theta$ -- line 9. \linebreak

At the end of the process, $f_\theta$ is a classifier able to properly predict the category of any image given any randomly sampled set of $B$ regions. As explained in the next section, the goal of learning the exploration policy is to improve the quality of $f_\theta$ by finding a good representation instead of using a uniform sampling approach.  This is done by finding a region selection policy that provides image-specific subsets of  $B$ image regions that are most likely to increase the classifier's classification accuracy.

\begin{algorithm}[t]
\begin{algorithmic}[1]
\REQUIRE $(x_1,...,x_\ell)$: Training set of images
\REQUIRE $(y_1,...,y_\ell)$: Training labels
\REQUIRE $B$: Budget
\REQUIRE $n$: 
\STATE $\mathcal{T}=\{\}$ \COMMENT{Training set}
\STATE \COMMENT{For each training image}
\FOR{$x_i$} 
\FOR{$k=1$ \TO $n$}
\STATE Sample $B$ regions $(s^{x_i}_1,...,s^{x_i}_B)$ using random exploration policy $(\pi_{random},...,\pi_{random})$
\STATE $\mathcal{T} \leftarrow \mathcal{T} \bigcup (\Phi(s^{x_i}_1,...,s^{x_i}_B),y_i)$
\ENDFOR 
\ENDFOR
\STATE Learn $f_\theta$ on $\mathcal{T}$ using a classical learning algorithm
\RETURN $f_\theta$
\end{algorithmic}
\caption{Classification Policy Learning Algorithm}
\label{alg3}
\end{algorithm}

\subsection{Learning the Optimal Exploration Policy}
\label{sec:ec}
Consider now that $f_\theta$ has been properly learned. Given a new image $x$, using only $f_\theta$ applied to a uniformly sampled set of regions has one main drawback: for some sampled sets, $f_\theta$ will certainly predict the right classification label, but for other samples, it will make a classification error, particularly for samples that contain irrelevant or misleading regions. The goal of the exploration policy $\pi$ is thus to provide $f_\theta$ with a set of \textit{good} regions i.e. a set of regions on which the classification function will predict the correct category. In other words, $\pi$ aims at reducing the error rate of $f_\theta$ by changing the way regions are sampled. The complete learning method is given in Algorithm \ref{alg4}.

\begin{figure}
\begin{center}
\begin{tabular}{p{1\linewidth}}
\includegraphics[width=1\linewidth]{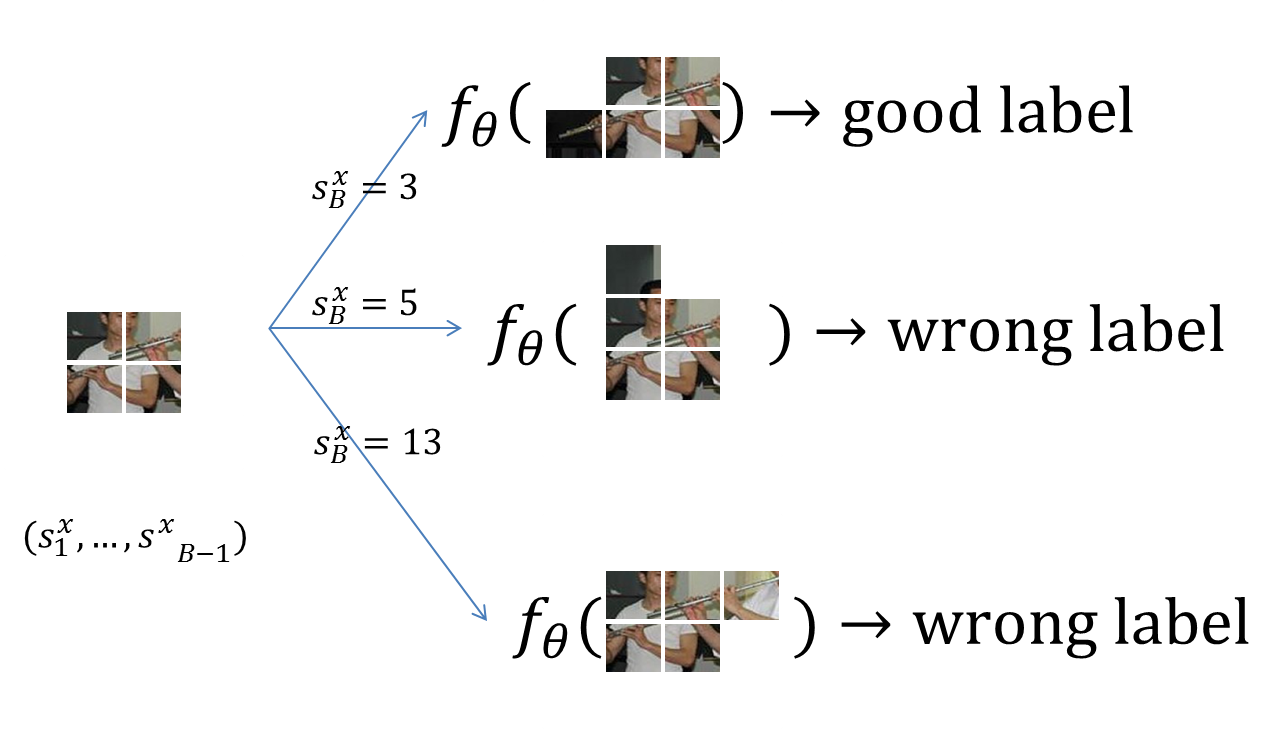} \\ 
\end{tabular}
\end{center}
\caption{Illustration of the learning algorithm for $\pi_{\gamma^{B-1}}$. On a sample $(s^x_1,...s^x_{B-1})$ of $B-1$ regions, all remaining regions $s^x_B$ are considered and classified by $f_\theta$ (left) (simulation step, lines 6-8 of Algorithm \ref{alg4}). For some regions -- the first one here -- $f_\theta$ computes the right label, for other regions $f_\theta$ makes a prediction error (line 9 of Alg. \ref{alg4}). The regions on which $f_\theta$ provides the good label are considered as training examples for learning  $\pi_{\gamma^{B-1}}$ (line 10).}
\label{fig:lfff}
\end{figure}

Given this principle, the idea of how to learn $\pi$ is as follows: consider the case where $f_\theta$ has been learned and we are currently learning $\pi_{\gamma^{B-1}}$ i.e. the sub-policy that aims at acquiring the $B$-th and final region. Given any sample of $B-1$ regions $(s^x_1,...s^x_{B-1})$, $\pi_{\gamma^{B-1}}$ can decide to acquire any of the remaining regions. If it acquires some of these regions, the classification policy will predict the right category while, for some other regions, $f_\theta$ will predict the incorrect category as illustrated in Fig.~\ref{fig:lfff}. The method we propose consists thus in simulating the decision of $f_\theta$ over all the possible $B$-th regions that can be acquired by $\pi_{\gamma^{B-1}}$, and to use the regions that correspond to a good final classification decision as supervised examples for learning $\pi_{\gamma^{B-1}}$. The result of this learning is to obtain a sub-policy that tends to select regions for which $f_\theta$ will be able to properly predict. \linebreak

The same type of reasoning can be also used for learning the other policies i.e. $\pi_{\gamma^{B-2}}$ will be learned in order to improve the quality of the sub-policy $(\pi_{\gamma^{B-1}}, f_\theta)$,  $\pi_{\gamma^{B-3}}$ will be learned in order to improve the quality of the sub-policy $(\pi_{\gamma^{B-2}}, \pi_{\gamma^{B-1}}, f_\theta)$ and so on. When learning sub-policy $\pi_{\gamma^t}$, we first start by building a simulation set where each element is an image represented by $t-1$ randomly sampled regions (line 5). Then, for each element, we test each possible remaining regions (lines 6-7) by simulating the previously learnt sub-policies $(\pi_{\gamma^{t+1}},...,f_\theta)$ (line 8). We thus build a training set (lines 9-10) that is used to learn $\pi_{\gamma^t}$ (line 15).

\begin{algorithm}[t]
\begin{algorithmic}[1]
\REQUIRE $(\pi_\gamma^{k+1},...,\pi_\gamma^{B-1},f_\theta)$: previously learned sub-policies
\REQUIRE $(x_1,...,x_\ell)$: Training set of images
\REQUIRE $(y_1,...,y_\ell)$: Training labels
\STATE $\mathcal{T}=\{\}$ \COMMENT{Training set}
\STATE \COMMENT{For each training image}
\FOR{$x_i$} 
\FOR{$k=1$ \TO $n$}
\STATE Sample $k-1$ regions $(s^{x_i}_1,...,s^{x_i}_{k})$ using random exploration policy
\STATE \COMMENT{For each region that has not been acquired}
\FOR{$s^{x_i}_{k+1} \notin s^{x_i}_1,...,s^{x_i}_{k}$} 
	\STATE Use sub-policies $(\pi_\gamma^{k+1},...,\pi_\gamma^{B-1},f_\theta)$ from $(s^{x_i}_1,...,s^{x_i}_{k},s^{x_i}_{k+1})$ to compute $\hat{y}$
	\IF{$\hat{y}==y_i$} \STATE $\mathcal{T} \leftarrow \mathcal{T} \bigcup (\Phi(s^{x_i}_1,...,s^{x_i}_{k}),s^{x_i}_{k+1})$ \ENDIF
\ENDFOR
\ENDFOR 
\ENDFOR
\STATE Learn $\pi_{\gamma}^{k}$ on $\mathcal{T}$ using a classical learning algorithm
\RETURN $\pi_{\gamma}^{k}$
\end{algorithmic}
\caption{Exploration Sub-policy $\pi_{\gamma^k}$  Learn. Algorithm}
\label{alg4}
\end{algorithm}

\subsection{Complexity}

The learning complexity of our method is the following: in order to simulate the behavior of the different sub-policies, we have to compute the features over all the regions of the training images. Moreover, for each training image, many samples of regions will be built (line 5 of Algorithms \ref{alg3} and \ref{alg4}). Let us denote $\ell$ the number of training images and $k$ the number of sample built for each image, at each step of the learning. The final complexity\footnote{We do not consider the complexity of the simulation phase of the algorithm i.e. line 8 of Algorithm \ref{alg4} which is usually negligible w.r.t the other factors of the learning method.} is $\mathcal{O}(N \times M \times C + B \times R(\ell \times k))$ where $R(n)$ corresponds to the complexity of learning over $n$ examples. $R$ depends on the machine learning algorithm used for representing $f_\theta$ and $\pi$. If we consider a classical BoW model, this complexity becomes $\mathcal{O}(N \times M \times C + R(\ell))$. Our method is thus slower to learn than standard models but still reasonable enough to allow the model to be learnt on large datasets.

\section{Experiments}
\label{sec:experiments}

We evaluate the proposed method on two challenging image databases corresponding to  different tasks: 
fine-grained image classification (People Playing Musical Instruments dataset\footnote{\href{http://ai.stanford.edu/~bangpeng/ppmi.html}\href{http://ai.stanford.edu/~bangpeng/ppmi.html}})~\cite{CVPR10_0463} and 
scene recognition  (15-scenes dataset)~\cite{lazebnik2006beyond}. 
Let us detail the experimental setup by presenting the low-level image features and the chosen vectorial representation for each region. We densely extract 
gray SIFT descriptors using the  VLFEAT library~\cite{vedaldi10vlfeat}. These local features are computed at a single scale ($s=16$ pixels) and with a constant step size $d$. 
For 15-scenes we use $d=8$ pixels while $d=4$ pixels for PPMI. Each region is represented by a BoW vector generated from SIFT descriptors~\cite{sivic03}. 
We run a K-Means algorithm by randomly sampling about 1 million descriptors in each database to produce a dictionnary of $M=200$ codeword elements. 
Each SIFT is then projected on the dictionary using hard assignement, and the codes are aggregated with sum pooling. The histogram is further $\ell_2$ normalized. 
Finally, we take the square root of each element thus generating a Bahttacharyya kernel feature map.\linebreak

The standard learning algorithm for $f_\theta$ and $\pi_\gamma$ is a one-against-all hinge-loss perceptron learned with a gradient descent algorithm. 
The gradient descent step and number of iterations have been tuned over the training set in order to maximize the accuracy of the classifier. 
We have chosen to create $10$ sequences of regions for each training images, resulting in a training set of size $10 \times \ell$ for each classifier -- $f_\theta$ and $\pi_{\gamma^t}$ -- learned by our method. 
Performance with more samples has been computed but is not reported here since it is equivalent to the one obtained with $10$ samples per image.

\subsection{Experimental Results}

In order to evaluate the performance of our method, we use the standard metrics for the two considered databases: multi-class accuracy for 15-scenes, and average accuracy over the 12 independently learned binary tasks for PPMI\footnote{Note that we do not report MAP metrics for PPMI, since our model produces a category label but not a score for each image.}.
We randomly sample training/testing images on $5$ splits of the data,  and the final performance corresponds to the average performance obtained on the 5 runs. 

We run experiments with different values of $B$. The baseline model performance is the one obtained where $B=16$ \textit{i.e.} all the regions are acquired by the model and the classification decision is based on the whole image. 
The results obtained with such a baseline approach are on par with previously published state-of-the art performances with a similar setup. 
For example,  we reach 77.7\% accuracy in the 15-Scene database, exactly matching the performances reported in~\cite{DBLP:conf/cvpr/PariziOF12}
\footnote{Their SBoW method matches our pipeline: mono-scale SIFT extracted with the same step size, same dictionary size, same coding/pooling schemes on a $4 \times 4$ grid,  and normalization policy.}. 
Although absolute performance can still be improved using more advanced low-level feature extraction or mid-level feature embedding, our main purpose here is to validate the relative performance advantage of the proposed method.
 
\begin{figure}
\begin{center}
\includegraphics[width=8.6cm]{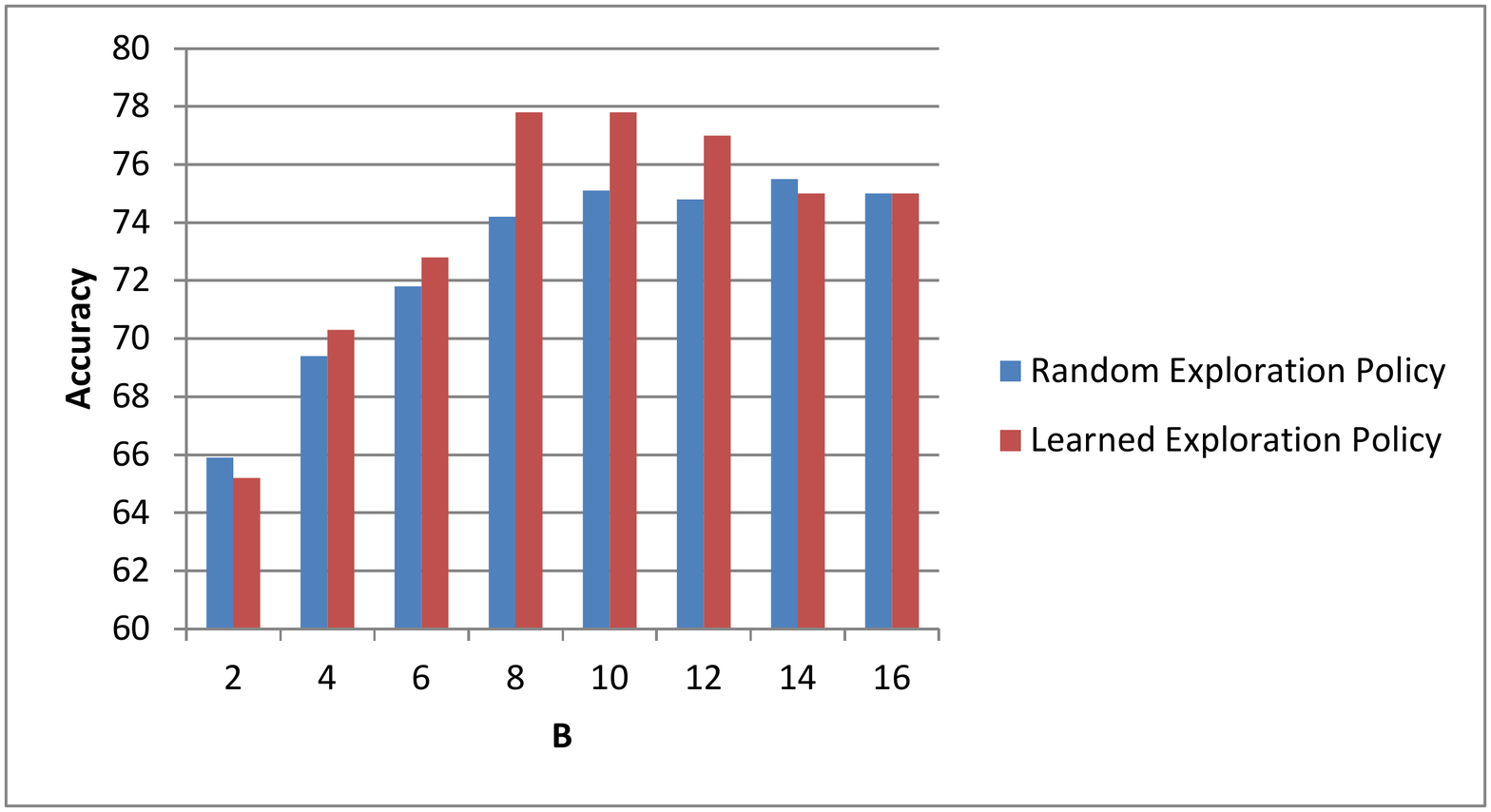}
\end{center}
\vspace{-1cm}
 \caption{Mean accuracy on the \textbf{PPMI dataset} depending on the budget, $B$.  We can see that the learned exploration policy (in red) has a better average accuracy for almost all budgets, especially when considering 8-12 regions.}
\label{r2}
\end{figure}

\begin{figure}
\begin{center}
\includegraphics[width=8.6cm]{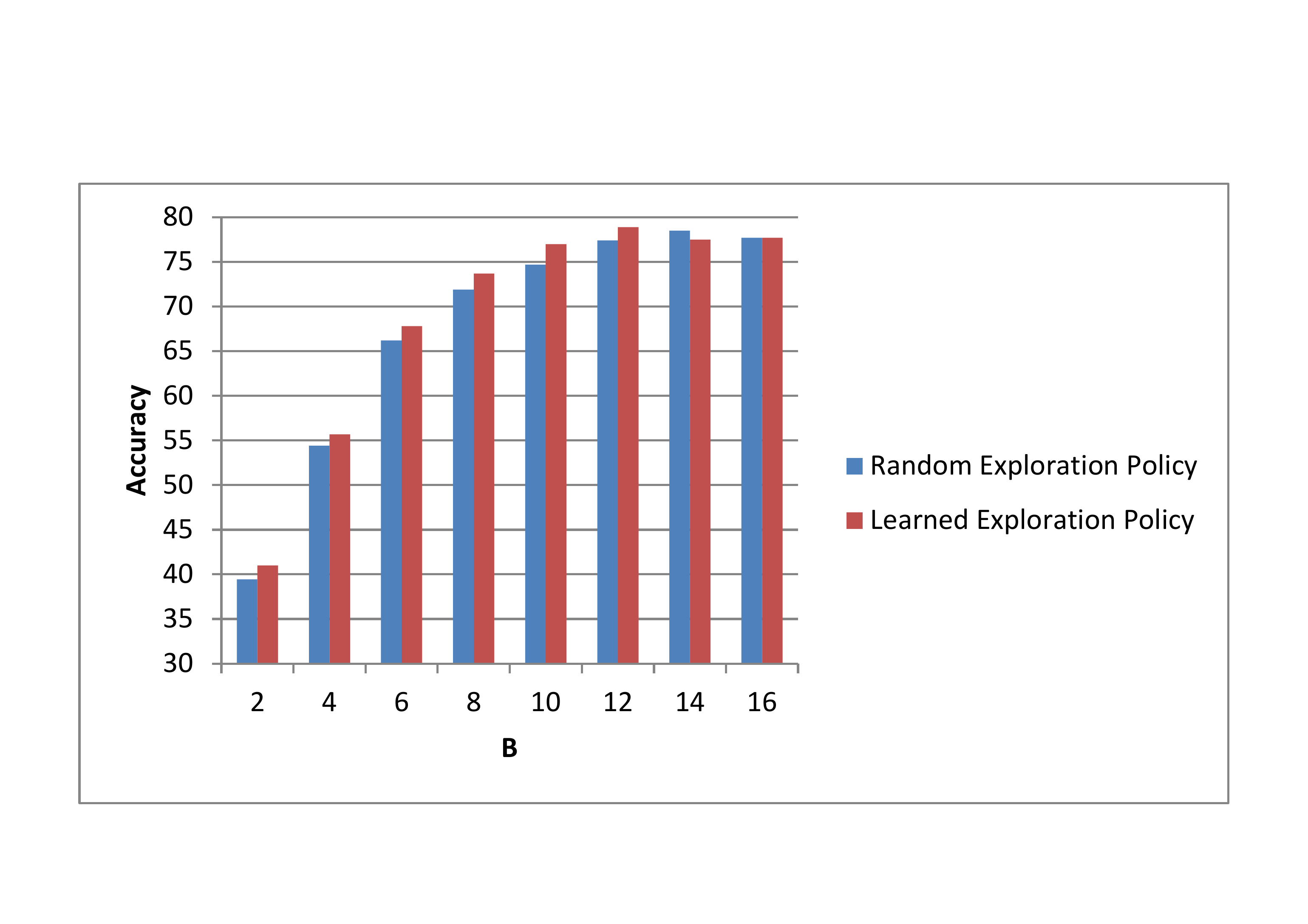}
\end{center}
\vspace{-1cm} 
\caption{Mean Accuracy on  \textbf{15 Scenes dataset} depending on the budget, $B$.  On this dataset, the learned exploration policy (red) is only slightly better than }
\label{r3}
\end{figure}

Figures \ref{r2} and \ref{r3} show the accuracy obtained for the the PPMI dataset and the 15 Scenes dataset. 
These figures present two measures: 
the performance obtained using a uniformly sampled subset of $B$ regions using $f_\theta$ --- in red ---, and the performance when the regions are sampled following the learned exploration policy $\pi_\gamma$ --- in blue.

\begin{figure}
\begin{center}
\includegraphics[width=1.0\linewidth]{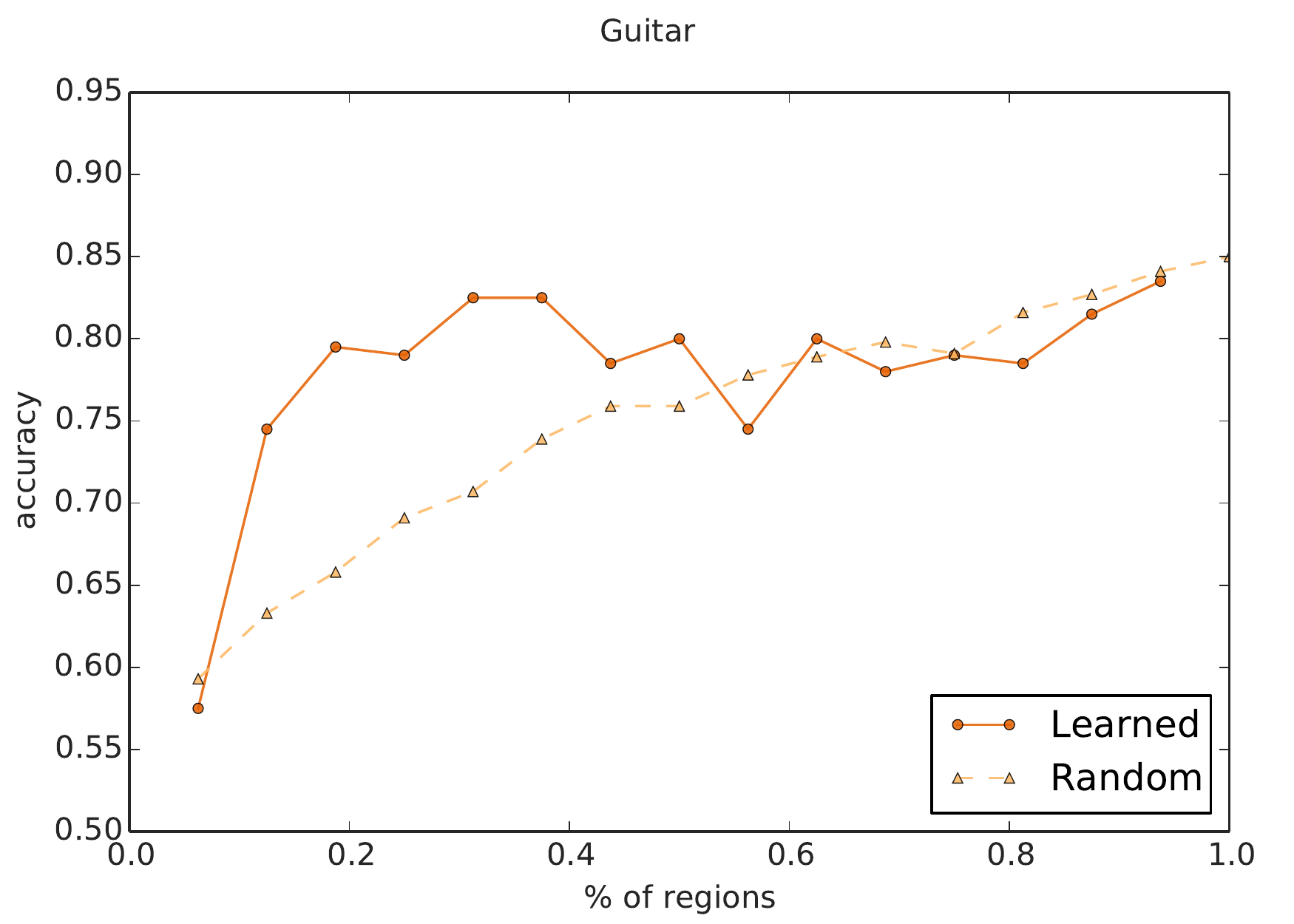}
\end{center}
\vspace{-0.5cm}
\caption{Accuracy on the Guitar dataset from PPMI with varying values of $B$.}
\vspace{-0.5cm}
\label{r1}
\end{figure}
\paragraph{Performance on a small set of regions vs performance of the baseline model: }
When comparing the accuracy of our method with $B < 16$ to the performance of the baseline method ($B=16$), one can see that given a reasonable value of $B$, our model is competitive with the classical approach. 
For example, acquiring  $B=10$ for the 15 Scenes dataset and for $B=8$ for the PPMI dataset allows one to obtain accuracy which is similar to the model with $B=16$. 
This means that our method is able to classify \textit{as well as} standard approach using only half of the regions. 
 Moreover, on PPMI, for $B=8, 10$ and $12$, our model clearly outperforms the baseline. 
This illustrates the ability of the learning algorithm to focus on relevant areas of the image, 
ignoring noisy or misleading regions. The generalization capacity of the prediction is thus  improved.  

\paragraph{Learned exploration policy vs random exploration policy: } 
Now, when comparing the performance of random exploration policy w.r.t. learned exploration policy, one can see that in almost all cases, the learned version is equivalent or better than the random one. For example, on the PPMI corpus with $B=8$, learning how to acquire the regions allow one to obtain an improvement of about $4\%$ in term of accuracy. This shows that our model is able, particularly on the PPMI dataset, to discover relevant regions depending on their content. This improvement is less important when the number of acquired regions is large. This is due to the fact that, when acquiring for example $B=12$ regions, even at random, the relevant information has a high chance of being acquired. The same effect also happens when $B$ is low e.g. when $B=2$, the information acquired is not sufficient to allow for good classification. This shows that our method is particularly interesting when the number of acquired regions is between about $30\%$ and $60 \%$ of the overall image.

On certain specific datasets performance gains with our method can be quit important.  Performance for varying values of $B$ for the Guitar dataset of PPMI is illustrated in Figure \ref{r1}, where we can see that as the percentage of regions acquired decreases (decreasing $B$), the random exploration policy's performance degrades quickly whereas our method is able to maintain adequate performance, until $B$ becomes too small.  The difference in performance between PPMI and 15 Scenes is likely linked to the fact that detecting instrument play in an image requires specific regions to be available (the face and the flute for example), whereas the scene can be inferred from a wider array of regions.

\begin{figure}[t]
\begin{center}
\includegraphics[width=1\linewidth]{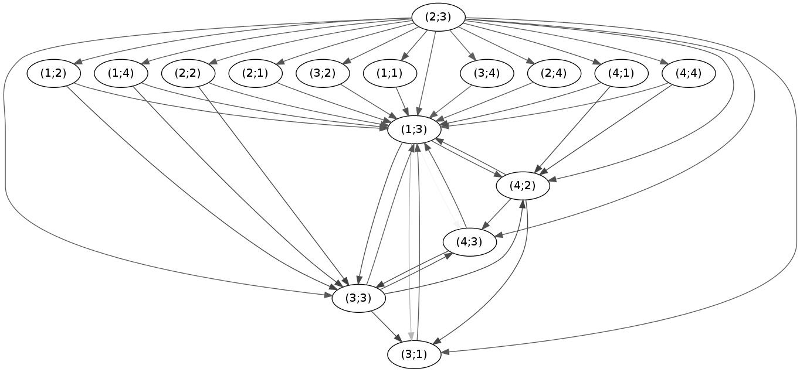}
\end{center}
\caption{Trajectories computed on PPMI-Flute (test). Each node corresponds to a region --- label is $(x,y)$-coordinates. Each edge $i\rightarrow j$ means that $j$ has been acquired just after having acquired $i$. The color of the edge represents the proportion of infered trajectories that contain the $i\rightarrow j$ transition.}
\label{f6}
\end{figure}

\paragraph*{Complexity Analysis:}
Let us denote $C$ the cost of computing the features over a region of the image, and $F$ the cost of computing $f_\theta$ or $\pi_{\gamma^i}$. The final inference complexity is $\mathcal{O}(B(C + F))$. 
As we use linear classifiers in all our experiments, classification time is insignificant in comparison to SIFT computation ($F<<C$), and complexity is therefore reduced to $\mathcal{O}(BC)$. In comparison to the cost of a classical BoW model $\mathcal{O}(NMB)$, the proposed model thus results in a speed-up of $\frac{N \times M}{B}$ .

\paragraph{Qualitative Results \& Analysis: } 

\begin{figure}
\begin{center}
\begin{tabular}{cc}
\includegraphics[width=.4\linewidth]{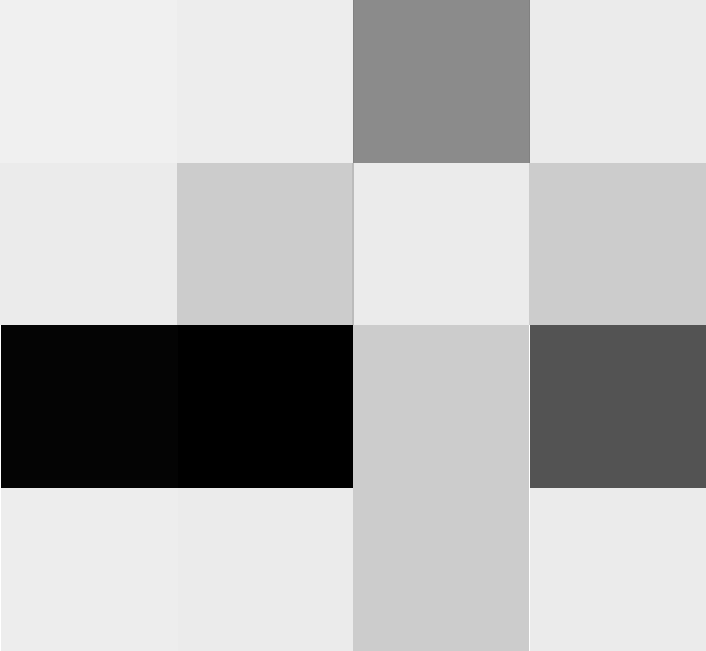} & \includegraphics[width=.4\linewidth]{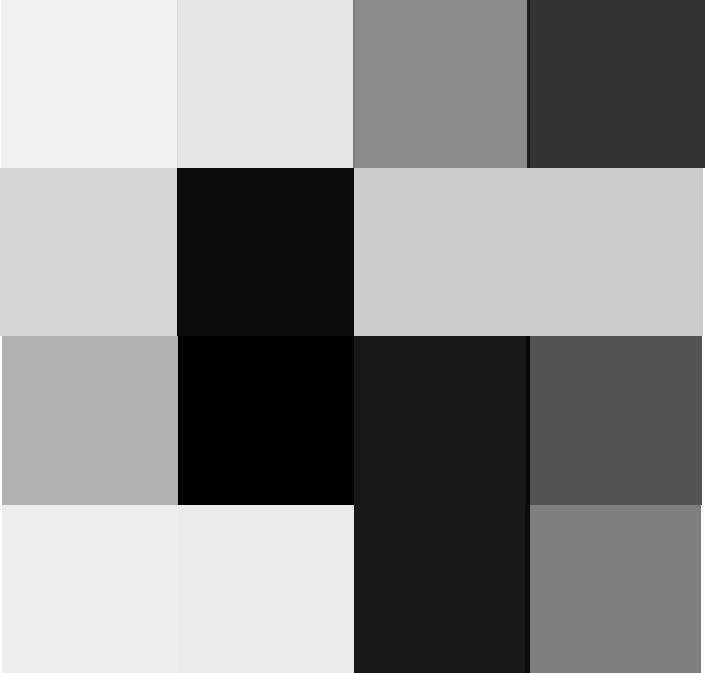} \\
\end{tabular}
\end{center}
\caption{Most frequent acquired regions for $B=4$ (left) and $B=8$ (right) on the PPMI Flute Dataset. The darker, the more frequent the region has been acquired for classifying.}
\label{f5}
\end{figure}

Figures \ref{f6} and \ref{f5} illustrate the learned exploration policy for the class flute of the PPMI dataset.
Figure \ref{f6} summarizes the regions visited over the testing images. 
Each region corresponds to a node of the graph, and an edge from region $i$ to region $j$ means that in at least one testing image, regions $j$ has been acquired immediately after region $i$. 
We can see that the algorithm is focusing its attention around 5 regions that are certainly relevant for many pictures, \textit{i.e.} $(1;3)$, $(4;2)$, $(4;3)$, $(3;3)$ and $(3;1)$. 
On the other hand, Figure \ref{f6} shows that at the beginning the algorithm tends to explore many different regions after the initial region.  
This shows that the the model starts by first exploring the image --- until having acquired 2 regions --- before focusing its attention on the regions that are the most relevant for predicting the category. 
Figure \ref{f5} shows the average behavior of our algorithm for $B=4$ and $B=8$.
We notice that about half of the acquired regions --- $2$ for $B=4$ and $4$ for $B=8$ --- are much more frequently explored than the other regions.
These regions certainly correspond to regions that generally carry relevant information on all the images. 


%

\begin{figure}
\begin{center}
\begin{tabular}{cc}
\includegraphics[width=.4\linewidth]{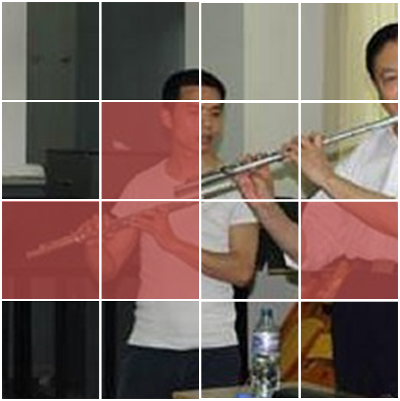} & \includegraphics[width=.4\linewidth]{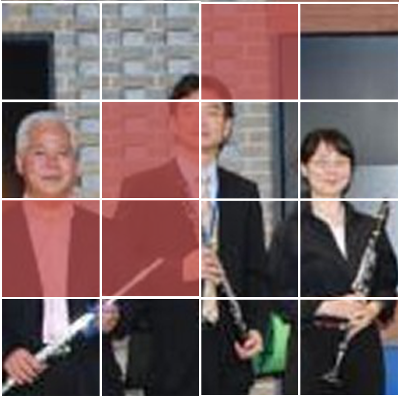} \\
\end{tabular}
\end{center}
\caption{Two examples of regions acquired with $B=4$ on the PPMI-Flute dataset. This example shows the ability of the model to adapt to different images where it is able to discover a flute.}
\label{f4}
\end{figure}

The images in Figure \ref{f5} can be interpreted as a spatial ``prior'' for a specific classification task. For example, for discriminating playing \textit{v.s.} holding flute with $B=2$, 
regions $(3;2)$ and $(3;1)$ are the more informative on average. 
However, since the decision is instance-based in our method, this spatial prior is balanced with the specific visual content of each test image. Our approach therefore shares some similarities with the reconfigurable model of \cite{DBLP:conf/cvpr/PariziOF12} using latent SVM. 
One example of instance-based classification is illustrated in figure \ref{f4}, where the regions visited with $B=4$ are shown. 
We can see that the set of regions visited changes between the left and right image. This illustrates the ability of our method to automatically adapt its choice of representation to the content of the image it is classifying. 

\section{Conclusion} \label{sec:conclusion}

In this paper, we introduced an adaptive representation process for image classification. The presented strategy combines both an exploration strategy used to find the best subset of regions for each image, and the final classification algorithm. New regions are iteratively selected based on the location and content of the previous ones. The resulting scheme produces an effective instance-based classification algorithm. We demonstrated the strategy's pertinence on two different image classification datasets. When using our exploration strategy limited to half of the regions of the images, we obtained a significant gain relative to baseline methods.
 


\bibliography{main}
\bibliographystyle{icml2014}

\end{document}